\documentclass[letterpaper]{article} 
\usepackage{aaai2026}  
\usepackage{times}  
\usepackage{helvet}  
\usepackage{courier}  
\usepackage[hyphens]{url}  
\usepackage{graphicx} 
\usepackage{natbib}  
\usepackage{caption} 
\usepackage{algorithm}
\usepackage{algorithmic}

\usepackage{newfloat}
\usepackage{listings}
\DeclareCaptionStyle{ruled}{labelfont=normalfont,labelsep=colon,strut=off} 
\floatstyle{ruled}
\newfloat{listing}{tb}{lst}{}
\floatname{listing}{Listing}
\title{Mind the Gap: Evaluating LLM Understanding of Human-Taught Road Safety Principles}
\author{
    Kranti Chalamalasetti\textsuperscript{\rm 1}\\
    kranti.chalamalasetti@uni-potsdam.de
}
\affiliations{
    \textsuperscript{\rm 1}University of Potsdam, Germany\\    

}

\usepackage{bibentry}

\begin{document}

\maketitle

\begin{abstract}
Following road safety norms is non-negotiable not only for humans but also for the AI systems that govern autonomous vehicles. In this work, we evaluate how well multi-modal large language models (LLMs) understand road safety concepts, specifically through schematic and illustrative representations. We curate a pilot dataset of images depicting traffic signs and road-safety norms sourced from school text books and use it to evaluate models capabilities in a zero-shot setting. 
Our preliminary results show that these models struggle with safety reasoning and reveal gaps between human learning and model interpretation. We further provide an analysis of these performance gaps for future research.
\end{abstract}


\section{Introduction}
Existing research on traffic and transportation systems largely focuses on issues such as congestion prediction~\citep{DBLP:journals/iotj/XuXQ25, DBLP:journals/tits/MahmudHAAAKS25, zhang2025mobile}, route optimization~\citep{huang2025multimodal, masri2025large}, or accident analysis~\citep{DBLP:journals/sensors/ZarzaCRC23a, DBLP:journals/corr/abs-2506-06301} that address the after-effects of unsafe conditions. However, far less attention has been given to the understanding of road safety, from the perspective of these AI-based systems. This understanding is also necessary for planners and architects to design safer urban environments.


Current AI models are typically trained on curated datasets such as real-world driving footage~\citep{Stallkamp-IJCNN-2011}, or traffic surveillance data~\citep{DBLP:journals/tits/MogelmoseTM12, Zhe_2016_CVPR}. However, these datasets capture the environment as it appears to machines, not as it is taught to humans. This raises a question: can AI models comprehend road safety concepts in the same way humans learn them? Humans begin learning these principles 
often through school textbooks that introduce road symbols, traffic signs, and safe behavioral norms such as \textit{cross the road at the zebra line} or \textit{red means stop before the line}. 

\begin{figure}[tb!]
  \includegraphics[width=0.48\textwidth]{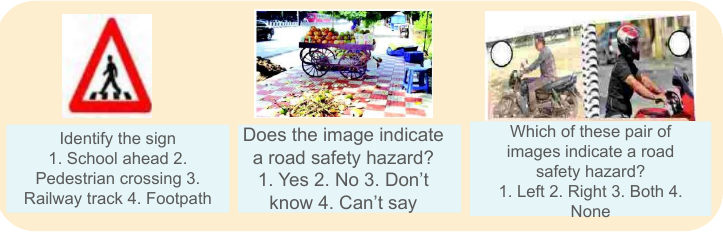}
  \caption{Road Safety Understanding Task Categories: Sign Detection, Road-Safety Hazard, and Hazard-Pair Comparison, each formatted as multiple-choice questions for zero-shot evaluation.}
  \label{fig:dataset_overview}
  \vspace*{-.4cm}
\end{figure}

In this paper, we propose a \textit{Road Safety Understanding Task} designed to evaluate how well multi-modal large language models comprehend basic road safety norms. The task is structured into three categories (see Figure~\ref{fig:dataset_overview}). Category 1 focuses on the detection and identification of traffic signs. Category 2 examines whether a given scenario adheres to or violates established safety norms. Category 3 presents a pair of images and asks the model to determine which one aligns correctly with road safety principles.

Since this task requires reasoning across both visual and textual modalities, it involves multiple sub-tasks: (a) the model must first interpret and understand the visual content of an image; (b) it must then map this information to known road safety norms; and (c) finally, it must answer a corresponding text-based question based on this interpretation.

Building on this motivation, our current work investigates the following research questions:
\begin{enumerate}\itemsep0em 
\item RQ1: To what extent do multimodal LLMs understand the representations of road safety concepts?

\item RQ2: How effectively can these models identify and reason about traffic signs, safe versus unsafe scenarios, and comparative visual cues related to road safety norms?

\item RQ3: What kinds of reasoning errors or conceptual gaps arise when models interpret human-taught representations of road safety?
\end{enumerate}

As a pilot study, we curated a dataset of 20-28 images per category. These images depict schematic scenes and symbols commonly used to teach road safety to children. 
We believe that the Road Safety Understanding Task is an initial step toward measuring and evaluating AI systems’ alignment with human safety education. By identifying gaps in model reasoning, we aim to guide the development of safety-aware intelligent transportation systems (ITS) and smarter, human-centered cities.

\section{Task and Dataset}
The Road Safety Understanding Task evaluates multimodal LLMs on their ability to reason about safety norms and behaviors by combining vision and language tasks to test if models can infer appropriate actions in given scenarios.

The task is framed as a question-answer problem. Each question consists of an image either a traffic sign or a visual scene, paired with a question that asks the model to identify the sign, determine whether the scene adheres to safety norms, or select the correct behavior. All categories contain multiple choice questions with four options, and the model is evaluated based on response accuracy. 
All images are 
resized to a 
target dimension (224×224) 
and padded to a square format with a white background to ensure uniform input size across models.

Motivated by recent efforts~\citep{DBLP:conf/nips/HuangBZZZSLLZLF23} to curate datasets from educational materials, we sourced images~\footnote{\url{https://github.com/kranti-up/urbansafetyliteracy/tree/main}} from Grade 7 textbooks published by the Andhra Pradesh State Board. We 
use schematic textbook illustrations because they represent the simplified, human-taught version of road safety concepts. These visuals abstract away the noise and variability of real-world imagery, allowing us to isolate the conceptual understanding of rules and behaviors.

\section{Findings and Discussion}

We evaluated three open-weight multimodal LLMs (Gemma-3-27B-IT, Qwen3-VL-30B, and InternVL3-78B), one closed model (GPT-5-Image-Mini), and a vision-only baseline (CLIP).
All multimodal models were accessed through the OpenRouter API~\footnote{\url{https://openrouter.ai/}} and queried in a zero-shot setting. Each prompt included the question, the image(s), and explicit instructions to answer with the corresponding multiple-choice option number. CLIP was loaded locally and evaluated using cosine-similarity scoring between image and text embeddings.

The results are summarized in Table~\ref{tab:prelimresults}. Overall, the multimodal LLMs performed well on Traffic-Sign Detection (SD) and Road-Safety Hazard (RSH) tasks but showed a drop in the Road-Safety Hazard-Pair (RSHP) task, which required comparative reasoning between two images. The CLIP baseline lagged behind in SD and RSH but achieved slightly higher relative accuracy in RSHP.

\paragraph{Traffic-Sign Detection (SD):} The 28 traffic sign samples were grouped~\footnote{\url{https://morth.nic.in/sites/default/files/road_safety_books.pdf}} into mandatory, cautionary, and informational types. Models achieved high accuracy on mandatory (e.g., Stop, One Way) and informational (e.g., Hospital, Petrol Pump) signs but only around 50\% on cautionary signs. Most errors were mirror-confusions for instance, misclassifying Right-hand Curve as Left-hand Curve or Right Hairpin Bend. This suggests that while the models have learned coarse sign semantics, they struggle with directional abstraction in textbook symbols.

\paragraph{Road-Safety Hazard (RSH):} In this task, a single image was presented and 
models had to determine whether the input image depicts a road-safety hazard (Yes / No / Don’t know / Can’t say). The multimodal LLMs performed well overall, but qualitative inspection revealed 
a few failure patterns: (a) context misinterpretation: in an image showing a child crossing the road while a green pedestrian signal was visible, three of the four models (all except InternVL) incorrectly classified it as a hazard, apparently assuming that a green signal implied vehicles would begin moving, while overlooking that the signal applied to pedestrians; (b) an over-cautious bias, where regular traffic scenes such as vehicles moving on a road or pedestrians walking on a footpath, are still labeled as hazards. Overall, the errors reveal that the models can identify visual cues associated with traffic but still lack a fine-grained understanding of safety contexts.

\paragraph{Road-Safety Hazard Pair (RSHP):} 

For this category, models frequently struggled with comparative reasoning. When a violation appeared in only one image, they often chose the opposite option, showing confusion with spatial references like ``left'' and ``right.'' Sometimes both images showed unsafe behavior but only one was selected, or both were correct yet the model still erred. These patterns suggest models respond to isolated cues rather than reasoning across scenes, leading to inconsistent judgments. The accuracy remains nearly unchanged when ``left/right image'' is replaced with ``first/second image,'' for most models, indicating that they largely rely on visual ordering rather than explicit spatial language grounding. InternVL model score improved to 0.24 from 0.14, suggesting partial sensitivity to linguistic framing of positional cues. 

For the CLIP baseline, performance varied noticeably with the phrasing of text embeddings. Using the exact label words (e.g., Stop, Give way) gave an accuracy of 0.39, which improved to 0.46 when ``sign'' was appended and further to 0.54 with ``symbol.'' A similar effect was observed in the RSH and RSHP tasks, where options such as yes/no or left/right were ineffective, but contextual phrases like “safe road behavior” and “traffic violation” significantly improved accuracy. These improvements show that CLIP’s alignment depends strongly on how the prompt matches the visual and linguistic domain of the data, with descriptive phrasing better reflecting the schematic style of textbook imagery.

\begin{table}[t]
    \centering
    \footnotesize
    \begin{tabular}{ccccc} \hline 
      Category   & SD (\#28) & RSH (\#20) & RSHP (\#21)  \\ \hline
      GPT5ImageMini & 0.79 & 0.80 & 0.14\\
      InternVL3-78B &  0.71 & 0.85 & 0.14 \\
      Qwen3-VL-30B & 0.68 & 0.65 & 0.10 \\
      Gemma3-27B-IT & 0.71 & 0.70 & 0.19 \\ \hline            
      CLIP Model & 0.54 & 0.35 & 0.38 \\ \hline                  
    \end{tabular}
    \caption{Zero-shot accuracy of vision–language models on three tasks: Sign Detection (SD, 28 samples), Road-Safety Hazard (RSH, 20 samples), and Hazard-Pair Comparison (RSHP, 21 samples)}
    \label{tab:prelimresults}
    \vspace*{-.4cm}
\end{table}

\section{Conclusion}
Overall, the results reveal that current multimodal LLMs can recognize urban symbols and scenes but struggle with schematic abstraction and comparative reasoning. Addressing these gaps is essential for developing AI systems that understand the visual language of urban safety and planning.

\section{Limitations}
This study has several limitations. The dataset is small and derived from a limited set of school textbooks, which may restrict generalizability across regions and visual styles. All evaluations were conducted in a zero-shot setting without model fine-tuning, providing diagnostic insights but not optimized performance. Finally, uniform image normalization, resizing and padding to square dimensions, may have introduced minor distortions or white-space artifacts that affect model sensitivity to layout and color cues.

\bibliography{aaai2026}

@article{DBLP:journals/iotj/XuXQ25,
  author       = {Zhengwei Xu and
                  Shaopeng Xu and
                  Zhihao Qu},
  title        = {Multilevel Spatial-Temporal Joint Large Language Model for Traffic
                  Prediction in Symbiotic IoT},
  journal      = {{IEEE} Internet Things J.},
  volume       = {12},
  number       = {20},
  pages        = {41390--41399},
  year         = {2025},
  url          = {https://doi.org/10.1109/JIOT.2025.3589247},
  doi          = {10.1109/JIOT.2025.3589247},
  bibsource    = {dblp computer science bibliography, https://dblp.org}
}

@article{DBLP:journals/tits/MahmudHAAAKS25,
  author       = {Doaa Mahmud and
                  Hadeel Hajmohamed and
                  Shamma Almentheri and
                  Shamma Alqaydi and
                  Lameya Aldhaheri and
                  Ruhul Amin Khalil and
                  Nasir Saeed},
  title        = {Integrating LLMs With {ITS:} Recent Advances, Potentials, Challenges,
                  and Future Directions},
  journal      = {{IEEE} Trans. Intell. Transp. Syst.},
  volume       = {26},
  number       = {5},
  pages        = {5674--5709},
  year         = {2025},
  url          = {https://doi.org/10.1109/TITS.2025.3528116},
  doi          = {10.1109/TITS.2025.3528116},
  bibsource    = {dblp computer science bibliography, https://dblp.org}
}

@article{zhang2025mobile,
  title={Mobile Traffic Prediction using LLMs with Efficient In-context Demonstration Selection},
  author={Zhang, Han and Sediq, Akram Bin and Afana, Ali and Erol-Kantarci, Melike},
  journal={IEEE Transactions on Communications},
  year={2025},
  publisher={IEEE}
}

@inproceedings{huang2025multimodal,
  title={How multimodal integration boost the performance of llm for optimization: Case study on capacitated vehicle routing problems},
  author={Huang, Yuxiao and Zhang, Wenjie and Feng, Liang and Wu, Xingyu and Tan, Kay Chen},
  booktitle={2025 IEEE Symposium for Multidisciplinary Computational Intelligence Incubators (MCII)},
  pages={1--7},
  year={2025},
  organization={IEEE}
}

@article{masri2025large,
  title={Large language models (llms) as traffic control systems at urban intersections: A new paradigm},
  author={Masri, Sari and Ashqar, Huthaifa I and Elhenawy, Mohammed},
  journal={Vehicles},
  volume={7},
  number={1},
  pages={11},
  year={2025},
  publisher={MDPI}
}

@article{DBLP:journals/sensors/ZarzaCRC23a,
  author       = {Irene de Zarz{\`{a}} and
                  Joachim de Curt{\`{o}} and
                  Gemma Roig and
                  Carlos T. Calafate},
  title        = {{LLM} Multimodal Traffic Accident Forecasting},
  journal      = {Sensors},
  volume       = {23},
  number       = {22},
  pages        = {9225},
  year         = {2023},
  url          = {https://doi.org/10.3390/s23229225},
  doi          = {10.3390/S23229225},
  bibsource    = {dblp computer science bibliography, https://dblp.org}
}

@article{DBLP:journals/corr/abs-2506-06301,
  author       = {Muhammad Monjurul Karim and
                  Yan Shi and
                  Shucheng Zhang and
                  Bingzhang Wang and
                  Mehrdad Nasri and
                  Yinhai Wang},
  title        = {Large Language Models and Their Applications in Roadway Safety and
                  Mobility Enhancement: {A} Comprehensive Review},
  journal      = {CoRR},
  volume       = {abs/2506.06301},
  year         = {2025},
  url          = {https://doi.org/10.48550/arXiv.2506.06301},
  doi          = {10.48550/ARXIV.2506.06301},
  eprinttype    = {arXiv},
  eprint       = {2506.06301},
  bibsource    = {dblp computer science bibliography, https://dblp.org}
}

@inproceedings{Stallkamp-IJCNN-2011,
    author = {Johannes Stallkamp and Marc Schlipsing and Jan Salmen and Christian Igel},
    booktitle = {IEEE International Joint Conference on Neural Networks},
    title = {The {G}erman {T}raffic {S}ign {R}ecognition {B}enchmark: A multi-class classification competition},
    year = {2011},
    pages = {1453--1460}
}

@article{DBLP:journals/tits/MogelmoseTM12,
  author       = {Andreas M{\o}gelmose and
                  Mohan M. Trivedi and
                  Thomas B. Moeslund},
  title        = {Vision-Based Traffic Sign Detection and Analysis for Intelligent Driver
                  Assistance Systems: Perspectives and Survey},
  journal      = {{IEEE} Trans. Intell. Transp. Syst.},
  volume       = {13},
  number       = {4},
  pages        = {1484--1497},
  year         = {2012},
  url          = {https://doi.org/10.1109/TITS.2012.2209421},
  doi          = {10.1109/TITS.2012.2209421},
  bibsource    = {dblp computer science bibliography, https://dblp.org}
}

@InProceedings{Zhe_2016_CVPR,

author = {Zhu, Zhe and Liang, Dun and Zhang, Songhai and Huang, Xiaolei and Li, Baoli and Hu, Shimin},

title = {Traffic-Sign Detection and Classification in the Wild},

booktitle = {The IEEE Conference on Computer Vision and Pattern Recognition (CVPR)},

year = {2016}

}

@Inproceedings{DBLP:conf/nips/HuangBZZZSLLZLF23,
  author       = {Yuzhen Huang and
                  Yuzhuo Bai and
                  Zhihao Zhu and
                  Junlei Zhang and
                  Jinghan Zhang and
                  Tangjun Su and
                  Junteng Liu and
                  Chuancheng Lv and
                  Yikai Zhang and
                  Jiayi Lei and
                  Yao Fu and
                  Maosong Sun and
                  Junxian He},
  editor       = {Alice Oh and
                  Tristan Naumann and
                  Amir Globerson and
                  Kate Saenko and
                  Moritz Hardt and
                  Sergey Levine},
  title        = {C-Eval: {A} Multi-Level Multi-Discipline Chinese Evaluation Suite
                  for Foundation Models},
  booktitle    = {Advances in Neural Information Processing Systems 36: Annual Conference
                  on Neural Information Processing Systems 2023, NeurIPS 2023, New Orleans,
                  LA, USA, December 10 - 16, 2023},
  year         = {2023},
  url          = {http://papers.nips.cc/paper\_files/paper/2023/hash/c6ec1844bec96d6d32ae95ae694e23d8-Abstract-Datasets\_and\_Benchmarks.html},
  bibsource    = {dblp computer science bibliography, https://dblp.org}
}


\end{document}